\documentclass[10pt,twocolumn,letterpaper]{article}
\usepackage[dvipsnames]{xcolor} 
\definecolor{green}{RGB}{0,128,0}
\definecolor{red}{RGB}{255,0,0}
\usepackage{booktabs, textcomp}
\usepackage{tabularx}   
\usepackage{booktabs}   
\usepackage{ragged2e}   
\usepackage[pagenumbers]{iccv} 
\usepackage[utf8]{inputenc}
\usepackage{amsmath,amssymb}  
\usepackage{algorithm}       
\usepackage{algorithmic}  
\usepackage{algorithm}
\usepackage{algorithmic}

\usepackage[utf8]{inputenc} 
\usepackage[T1]{fontenc}    
\usepackage{url}            
\usepackage{booktabs}       
\usepackage{amsfonts}       
\usepackage{nicefrac}       
\usepackage{microtype}      
\usepackage{amsthm}
\usepackage{amsmath}
\usepackage{graphicx}

\usepackage{bbm}
\usepackage{diagbox}
\usepackage{colortbl}
\usepackage{multirow}
\usepackage{pifont} 
\usepackage{bm}
\usepackage{subcaption}
\usepackage{multirow}
\usepackage{threeparttable}
\usepackage{multirow}
\usepackage[export]{adjustbox}
\usepackage{algorithm}
\usepackage{algorithmic}
\usepackage{mdframed}

\usepackage{amsmath}    
\usepackage{amssymb}    
\usepackage{amsfonts}   

\usepackage{graphicx}   
\usepackage{float}

\usepackage{makecell}
\usepackage{xcolor} 
\definecolor{snowblue}{rgb}{0.65, 0.80, 0.98} 
\def\method{SDKD}
\definecolor{iccvblue}{rgb}{0.21,0.49,0.74}
\usepackage[pagebackref,breaklinks,colorlinks,allcolors=iccvblue]{hyperref}

\def\paperID{12201} 
\def\confName{ICCV}
\def\confYear{2025}

\title{Frequency-Aligned Knowledge Distillation for Lightweight \\ Spatiotemporal Forecasting}

\author{
Yuqi Li\textsuperscript{1}\thanks{Equal Contribution.} \qquad 
Chuanguang Yang\textsuperscript{1}\footnotemark[1] \qquad
Hansheng Zeng\textsuperscript{2} \qquad 
Zeyu Dong\textsuperscript{1},\\
Zhulin An\textsuperscript{1} \qquad
Yongjun Xu\textsuperscript{1} \qquad
Yingli Tian\textsuperscript{3} \qquad
Hao Wu\textsuperscript{4}\thanks{Corresponding Author.}
 \\
\textsuperscript{1}Institute of Computing Technology, Chinese Academy of Sciences \\
\textsuperscript{2}The University of Hong Kong\\
\textsuperscript{3}The City University of New York\\
\textsuperscript{4}Tsinghua University\\
}

\begin{document}
\maketitle

\begin{abstract}
Spatiotemporal forecasting tasks, such as traffic flow, combustion dynamics, and weather forecasting, often require  complex models that suffer from low training efficiency and high memory consumption. This paper proposes a lightweight framework, \textbf{S}pectral \textbf{D}ecoupled \textbf{K}nowledge \textbf{D}istillation (\textit{termed~\method{}}), which transfers the multi-scale spatiotemporal representations from a complex teacher model to a more efficient lightweight student network. The teacher model follows an encoder-latent evolution-decoder architecture, where its latent evolution module decouples high-frequency details and low-frequency trends using convolution and Transformer (global low-frequency modeler). However, the multi-layer convolution and deconvolution structures result in slow training and high memory usage. To address these issues, we propose a frequency-aligned knowledge distillation strategy, which extracts multi-scale spectral features from the teacher’s latent space, including both high and low frequency components, to guide the lightweight student model in capturing both local fine-grained variations and global evolution patterns. Experimental results show that \method{} significantly improves performance, achieving reductions of up to 81.3\% in MSE and in MAE 52.3\% on the Navier-Stokes equation dataset. The framework effectively captures both high-frequency variations and long-term trends while reducing computational complexity. Our codes are available at \url{https://github.com/itsnotacie/SDKD}
\end{abstract}

\vspace{-15pt}
\section{Introduction}
Spatiotemporal forecasting, such as traffic flow~\cite{jiang2022graph, wu2025turb, wu2024earthfarsser, pan2019urban,yu2025ginar+, wu2024pure} prediction and weather evolution modeling~\cite{bi2023accurate,shu2025ocean, zhang2023skilful,yu2025merlin}, is a core task for real-time decision-making in smart cities and climate science~\cite{milojevic2021machine}. These tasks require modeling the complex coupling between high-frequency local patterns (e.g., sudden traffic congestion)~\cite{dixon2019deep, gao2025neuralom} and low-frequency global evolution (e.g., slow atmospheric pressure changes)~\cite{simmons2014estimating}. Recently, hybrid architectures combining convolutional neural networks (CNNs) and Transformers have achieved state-of-the-art forecasting accuracy by capturing both local features and long-range dependencies~\cite{wu2024pastnet, wu2024earthfarsser}. However, as shown in Table~\ref{tab:complexity}, their deep stacked structures and complex operators (e.g., multi-head attention with quadratic complexity $O(N^2d)$) lead to prohibitive computational costs and memory consumption (e.g., $3d^2$ parameters per attention layer), making them difficult to deploy in resource-constrained environments such as edge devices. Although model compression techniques like knowledge distillation (KD) offer potential solutions, existing KD frameworks—mainly designed for classification tasks~\cite{yang2021hierarchical,yang2022mutual,yang2023online}—perform poorly when directly applied to spatiotemporal forecasting. The fundamental reason lies in the multi-band spectral characteristics of continuous spatiotemporal signals, where simple feature imitation fails to maintain the critical balance between high-frequency details and low-frequency trends.

Traditional approaches attempt to decouple spatial and temporal modeling through inductive bias of CNNs/RNNs (e.g., ConvLSTM~\cite{shi2015convolutional} for localized spatial correlations and PredRNN~\cite{wang2017predrnn} for sequential dynamics). However, these methods inherently struggle to model cross-scale interactions—the intricate coupling between high-frequency transients (e.g., abrupt traffic surges) and low-frequency trends (e.g., diurnal periodicity). While hybrid CNN-Transformer architectures~\cite{wu2024earthfarsser, wu2025advanced} mitigate this by jointly capturing local-global dependencies, their computational complexity scales quadratically with sequence length (see Table~\ref{tab:complexity}), rendering them impractical for long-horizon forecasting. Prior knowledge distillation (KD) methods for regression tasks, such as feature mimicry~\cite{romero2014fitnets} or output distribution alignment~\cite{yu2024decoupling}, further exacerbate this issue by ignoring the spectral bias inherent in spatiotemporal signals. For instance, distilling high-level semantic features disproportionately suppresses low-frequency components critical for long-term consistency~\cite{lin2022investigating}, while naive parameter reduction in lightweight architectures (e.g., depthwise separable convolutions~\cite{sarkar2021understanding}) sacrifices generalizability across heterogeneous domains (e.g., fluid dynamics vs. urban mobility). \textbf{\textit{This dual challenge of efficiency-accuracy trade-off and spectral imbalance underscores the need for a principled distillation framework that explicitly preserves multi-scale spatiotemporal patterns.}}

\noindent\ding{224} \textbf{\textit{Key Gap}.} Existing approaches suffer from a fundamental spectral entanglement dilemma: \textit{the absence of explicit mechanisms to decouple and transfer frequency-specific knowledge across spatiotemporal scales.} This leads to a paradoxical trade-off in lightweight models—either over-smoothing high-frequency variations (e.g., traffic surge spikes) through excessive low-pass filtering~\cite{rahaman2019spectral}, or failing to capture slow-evolving trends (e.g., seasonal climate shifts) due to high-frequency dominance in parameter updates~\cite{cao2019towards}. While recent attempts like frequency-aware KD~\cite{zhang2024freekd} apply static spectral masks for classification tasks, they ignore the non-stationary spectral properties of spatiotemporal dynamics, where optimal frequency bands evolve both spatially (e.g., urban vs. rural regions) and temporally (e.g., peak vs. off-peak hours).

We propose \textbf{SDKD} (Spectral Decoupled Knowledge Distillation), a principled framework that bridges the efficiency-accuracy gap by compressing complex spatiotemporal teachers into deployable lightweight students through frequency-aware representation alignment. Our core insight stems from the spectral duality of spatiotemporal signals: they inherently comprise \textit{high-frequency components} (localized rapid variations, e.g., traffic congestion spikes) governed by spatial locality, and \textit{low-frequency components} (global slow dynamics, e.g., diurnal traffic periodicity) dominated by temporal continuity~\cite{li2020fourier,cao2019towards}. SDKD exploits this duality through two synergistic innovations:
\begin{itemize}[leftmargin=*]
\item[\ding{224}] \textbf{\textit{Teacher Design with Spectral Disentanglement}.} We architect the teacher model as a frequency-aware autoencoder, where CNNs act as \textit{high-pass filters} to capture fine-grained spatial patterns through localized gradient operations~\cite{wang2020high, abello2021dissecting}, while Transformers serve as \textit{low-pass filters} to model global temporal trends via attention-based trend decomposition~\cite{raghu2021vision, park2022vision}. This explicit frequency decoupling in the latent space creates disentangled spectral priors—a critical foundation for effective distillation.

\item [\ding{224}] \textbf{\textit{Architecture-Independent Spectral Transfer}.} Traditional knowledge distillation methods rely on specific architectures to transfer knowledge. However, in spatio-temporal prediction tasks, different student models (e.g., U-Net, ResNet) have different structural characteristics and handle spectral information in various ways. Therefore, we propose an architecture-independent spectral transfer mechanism, which focuses on directly aligning spectral features for knowledge transfer, rather than on the student model's structure.

\end{itemize}

\noindent\textbf{\textit{Contributions Summary.}} This paper proposes a systematic solution to balance efficiency and accuracy while addressing spectral imbalance in spatiotemporal forecasting. The main contributions are as follows:

\begin{itemize}[leftmargin=*]
\item[\ding{224}] \textbf{\textit{Spectral-Decoupled Teacher Architecture}}. This paper introduces a frequency-unraveling teacher model with a convolution-Transformer serial design. It explicitly separates high-frequency details (local gradient response) from low-frequency trends (global attention modeling) to construct interpretable spectral priors. Theoretical analysis~\cite{park2022vision} shows that this design aligns the spectral energy distribution in the latent space with physical laws. 

\item[\ding{224}] \textbf{\textit{Frequency-Aligned Distillation Mechanism}}. This paper designs a cross-scale spectral feature alignment loss. It uses multi-band components from the teacher’s latent space as dynamic supervision signals to guide the student model in capturing transient fluctuations and evolutionary trends. This mechanism applies spectrum-sensitive adaptive weighting to mitigate the student model's inherent high-frequency overfitting and low-frequency underfitting.

\item[\ding{224}] \textbf{\textit{Excellent Performance}}. Experiments show that ST-AlterNet performs better than SimVP, especially on the Weatherbench dataset, with higher MAE, PSNR, and SSIM scores. Using the AEKD strategy (especially AB Loss or MSE Loss) significantly improves student models such as U-Net, ResNet, and MLP-Mixer in MAE and SSIM. For example, U-Net with AEKD (AB Loss) achieves MAE=0.8541, PSNR=31.4527, and SSIM=0.8812 on Weatherbench, showing the effectiveness of AEKD in improving student performance.

\end{itemize}
\noindent This framework provides a general paradigm for high-accuracy spatiotemporal forecasting in resource-limited scenarios.

\begin{table}[h]
  \caption{Model Complexity and Parameter Count (\textit{N}: sequence length, \textit{d}: channels, \textit{K}: kernel size, $|E|$: edges)}
  \small
  \centering
  \resizebox{\columnwidth}{!}{
    \begin{tabular}{lccc}
      \toprule
      \textbf{Model} & \textbf{FLOPs} & \textbf{Parameters} & \textbf{Description} \\
      \midrule
      CNN & $K^2d^2N$ & $K^2d^2+d$ & Local filtering \\
      RNN (LSTM) & $8d^2N$ & $4d^2+4d$ & Temporal modeling \\
      Self-Attention & $N^2d+3Nd^2$ & $3d^2$ & Global context \\
      GNN & $|E|d+Nd^2$ & $d^2+d$ & Graph propagation \\
      FNO & $Nd^2+Nd\log N$ & $Nd^2$ & Spectral conv. \\
      \bottomrule
    \end{tabular}
  }
  \label{tab:complexity}
  \vspace{-0.5cm}
  
\end{table}

\section{Related Work}

\noindent\textbf{Spatiotemporal Forecasting Models.} The core challenge in spatiotemporal forecasting is modeling the complex coupling between local spatial patterns and global temporal evolution in dynamic systems. Early works optimized efficiency by separating spatial and temporal modeling. ConvLSTM~\cite{shi2015convolutional} introduced convolution into LSTM to capture spatial correlations, while PredRNN~\cite{wang2017predrnn} stacked memory units to enhance temporal modeling. However, these methods struggle with cross-scale interactions, such as synchronizing high-frequency transient fluctuations and low-frequency trends. Recently, hybrid architectures (e.g., CNN-Transformer) improved forecasting accuracy by integrating local convolution and global attention mechanisms~\cite{wu2024earthfarsser}, but their quadratic computational complexity and high memory cost limit practical deployment. To improve efficiency, HPF~\cite{pham2024frequency} introduced a lightweight temporal modeling module but sacrificed the ability to capture high-frequency details. Unlike these approaches, this paper reconstructs the modeling paradigm from a frequency-domain decoupling perspective, using knowledge distillation to balance accuracy and efficiency.

\noindent\textbf{Knowledge Distillation Techniques.} Knowledge distillation (KD)~\cite{yang2022cross,li2023curriculum,yang2024clip,li2024promptkd} compresses models by transferring knowledge from a complex teacher model to a lightweight student model. Traditional KD methods focus on classification tasks, transferring knowledge through softened label distributions~\cite{hinton2015distilling} or intermediate feature matching~\cite{ji2021show}. For regression tasks, FitNet-R~\cite{romero2014fitnets} aligns multi-scale features, but directly applying it to spatiotemporal forecasting often causes high-frequency information loss due to spectral bias. Recent studies incorporate frequency-domain analysis to improve distillation. FOLK~\cite{karimi2024frequency} separates high and low frequency knowledge using frequency masks in image classification but ignores the continuity of spatiotemporal signals. STGNN-Distill~\cite{izadi2024knowledge} enhances long-range dependency modeling through spatiotemporal attention distillation but still relies on complex attention mechanisms. This paper introduces the first decoupled distillation framework tailored for spatiotemporal signals, enabling cross-scale knowledge transfer through a frequency-domain alignment strategy.

\noindent\textbf{Frequency-Domain Representation Learning.} Frequency domain analysis provides a theoretical tool for understanding the spectral preferences of deep networks. Jacot et al.~\cite{jacot2018neural} showed that CNNs exhibit an inductive bias toward low-frequency components through neural tangent kernel theory, while Transformers, due to their self-attention mechanism, capture high-frequency details more effectively~\cite{cao2021choose}. Following this, FNO~\cite{li2020fourier} models partial differential equations using Fourier operators but struggles with non-stationary spatiotemporal signals. SFNet~\cite{cui2023selective} decouples multi-scale features using wavelet transforms, but its manually designed filters lack adaptability. Unlike explicit frequency-domain modeling, this paper leverages a teacher model with implicit spectral decoupling to provide prior knowledge, allowing the student model to inherit multi-scale representations without complex frequency-domain operations. This approach introduces a new paradigm for lightweight spatiotemporal forecasting.

\noindent \textbf{\textit{Key Differences}.} Existing works either rely on complex hybrid architectures that sacrifice efficiency (e.g., CNN-Transformer), simplify models at the cost of fine-grained dynamics (e.g., FOLK), or use generic distillation strategies unrelated to the task (e.g., FitNet-R).  
This paper introduces three key innovations:  
1) A frequency-domain decoupled teacher model that explicitly separates multi-scale spatiotemporal patterns.  
2) A spectral-aligned distillation mechanism that mitigates cross-scale modeling biases in lightweight models.  
3) A theoretical proof demonstrating the necessity of spectral knowledge transfer for learning long-range dependencies, providing an interpretable approach for lightweight spatiotemporal forecasting.
\section{Method}
\begin{figure*}[t]
\centering
\includegraphics[width=1\textwidth]{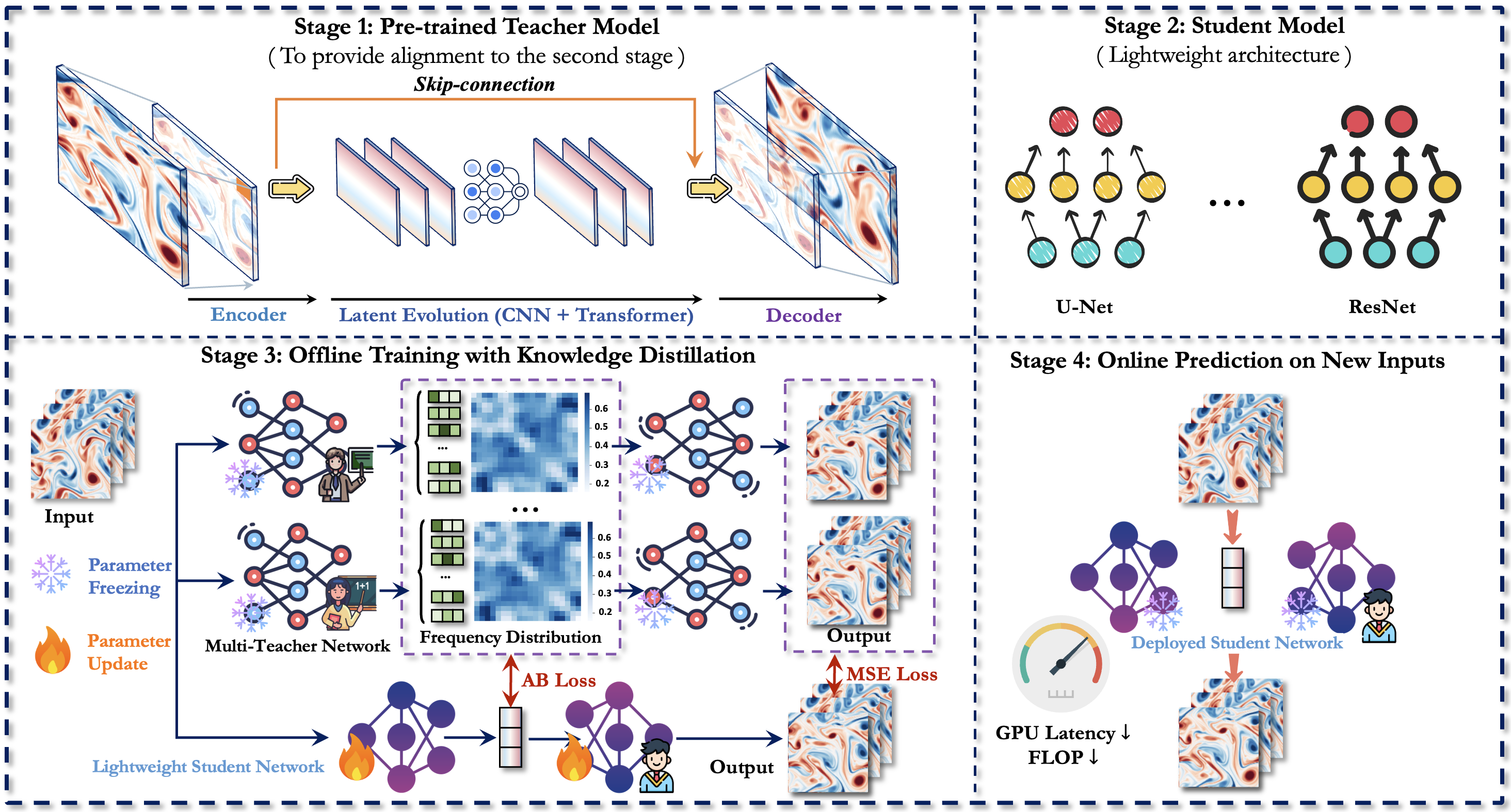}
\caption{Architecture Overview of~\method{}.
\textbf{Stage 1}: Teacher model pretraining with frequency decoupling.
\textbf{Stage 2}: Lightweight student architecture construction using parameter-efficient backbones like ResNet or U-Net.
\textbf{Stage 3}: Offline Training with Knowledge Distillation.
\textbf{Stage 4}: Online Prediction on New Inputs.}
\label{fig:KD_main} 
\vspace{-0.3cm}
\end{figure*}

\subsection{Problem Formulation}
Given a spatiotemporal sequence $X = \{X_t \in \mathbb{R}^{H \times W \times C}\}_{t=1}^{T}$, with $H \times W$ spatial grids and $C$ channels, the forecasting task aims to predict future states $Y = \{X_{T+\tau}\}_{\tau=1}^{\Delta}.$ Traditional methods minimize 
$\|F(X) - Y\|$, where $F$ is a hybrid CNN-Transformer model. However, the quadratic complexity of Transformers $\mathcal{O}(N^2 d)$ and deep CNNs limits deployment efficiency. Our goal is to distill $F$ into a lightweight student $\mathcal{G}$ with:
\begin{equation}
\min_{\theta_{\mathcal{G}}} 
   \underbrace{\bigl|\mathcal{G}(X) - Y\bigr|}_{\text{Forecasting Loss}}
   \;+\; 
   \lambda \,
   \underbrace{\mathcal{D}\!\Bigl(\Psi\bigl(\mathcal{F}(X)\bigr), \,\Psi\bigl(\mathcal{G}(X)\bigr)\Bigr)}_{\text{Spectral Distillation Loss}}.
\label{eq:distillation}
\end{equation}
Here, $\Psi$ extracts spectral features, and $\lambda$ balances accuracy and efficiency.

\subsection{Stage 1: Teacher Model Pretraining: Implicit Spectrum Decoupling}
The teacher model $\mathcal{F}$ uses the \textbf{Encoder-Latent Evolution-Decoder} architecture in Figure~\ref{fig:KD_main}. It decouples frequency components in the latent space by cascading convolution and Transformer. We formalize it as follows:

\begin{equation}
Z = \mathcal{E}(X) \in \mathbb{R}^{H' \times W' \times D}.
\end{equation}

The encoder $\mathcal{E}$ maps the input sequence $X$ to the latent space $Z$ through multiple convolution layers. Then the latent evolution module processes $Z$ with two parallel frequency-sensitive operations:

\paragraph{High-Frequency Extractor (CNN as High-Pass Filter)}
\begin{equation}
Z^h = \text{ConvBlock}(Z) = \sigma(\text{Conv2D}(Z; \mathbf{W}_h) + b_h).
\end{equation}

Here $\mathbf{W}_h \in \mathbb{R}^{K\times K\times D\times D}$ is the convolution kernel, and $K$ is the local receptive field size. According to Park et al.~\cite{park2022vision}, convolution layers perform implicit high-frequency filtering through local gradient operators $\nabla_{x,y}$:

\begin{equation}
\mathcal{F}_{\text{high}}(Z)(x,y) = \sum_{|\alpha|\leq k} a_\alpha \cdot \partial^\alpha Z(x,y).
\end{equation}

The term $\alpha$ is the differentiation order, and $a_\alpha$ is a learnable parameter. This property helps CNN to effectively capture high-frequency details like abrupt traffic changes and turbulent vortices.

\paragraph{Low-Frequency Modeler (Transformer as Low-Pass Filter)}
\begin{equation}
Z^l = \text{TransformerBlock}(Z) = \text{Softmax}\left(\frac{QK^T}{\sqrt{d}}\right)V.
\end{equation}

Here $Q,K,V \in \mathbb{R}^{N \times d}$ are the query, key, and value matrices in self-attention, and $N=H' \times W'$ is the number of spatial positions. According to neural tangent kernel theory\cite{jacot2018neural}, self-attention acts like a low-pass filter:

\begin{equation}
\mathcal{F}_{\text{low}}(Z)(\omega) = \frac{1}{1 + \|\omega\|^2/\lambda} \cdot \hat{Z}(\omega).
\end{equation}

The variable $\omega$ is the frequency component, and $\lambda$ is the smoothing coefficient. This property helps the Transformer model low-frequency trends in meteorological evolution.

\paragraph{Latent State Fusion}
We fuse the high- and low-frequency components by a residual connection:
\begin{equation}
Z_{\text{evolved}} = \text{LayerNorm}(Z^h + Z^l).
\end{equation}

Finally, the decoder $\mathcal{D}$ reconstructs the prediction via deconvolution:
\begin{equation}
\hat{Y} = \mathcal{D}(Z_{\text{evolved}}).
\end{equation}

This design ensures frequency decoupling in the latent space. According to Plancherel’s theorem, the total error decomposes as:
\begin{equation}
\begin{split}
\|\hat{Y}-Y\|^2 
= 
& \underbrace{\sum_{|\omega|>\omega_c} |\hat{Y}^h(\omega)-Y(\omega)|^2}_{\text{High-Frequency Error}} \\
& \hspace{0.5em} + \underbrace{\sum_{|\omega|\leq\omega_c} |\hat{Y}^l(\omega)-Y(\omega)|^2}_{\text{Low-Frequency Error}}.
\end{split}
\end{equation}

Here $\omega_c$ is the cutoff frequency. By alternating the optimization of CNN and Transformer modules, the teacher model minimizes both high- and low-frequency errors and provides ideal spectral priors for knowledge distillation.

\noindent\textbf{Relation to Physical Laws:} This architecture draws inspiration from the Navier-Stokes equations~\cite{li2020fourier}. The high-frequency component corresponds to the viscous dissipation term $\nu\nabla^2\mathbf{u}$, and the low-frequency component corresponds to the convection term $(\mathbf{u}\cdot\nabla)\mathbf{u}$. Experimental results (see Figure~\ref{fig:kd_low_high}) show that the latent space aligns with the Kolmogorov energy spectrum $E(\omega) \propto \omega^{-5/3}$ in fluid dynamics, which confirms its physical interpretability.

\subsection{Stage 2: Lightweight Student Architecture}
To enable efficient deployment, we design a lightweight student model $\mathcal{G}$ that can be flexibly instantiated as a simple CNN (e.g., ResNet), a U-Net, or even a multi-layer perceptron (MLP) with spatial reshaping. The key design goal is to avoid the heavy Transformer blocks and deep convolution stacks in the teacher model, drastically reducing parameter count and training memory while still maintaining adequate capacity for multi-scale spatiotemporal patterns.

\paragraph{Model Structure.} Figure~\ref{fig:KD_main} (Stage 2) shows a compact variant that replaces the teacher’s complex latent evolution module with a shallower backbone, such as:
\begin{itemize}[leftmargin=*]
\item \textbf{ResNet}~\cite{he2016deep}: A few residual blocks (3--4 layers each) for local feature extraction, enhanced by skip connections that help preserve gradient flow.  
\item \textbf{U-Net}~\cite{ronneberger2015u}: An encoder-decoder with symmetric skip connections across scales to capture both coarse and fine spatial features.  
\item \textbf{MLP}~\cite{tolstikhin2021mlp}: Flatten (or reshape) the spatiotemporal grid and apply linear layers, suitable for extremely low-resource settings.  
\end{itemize}

These backbones keep computational overhead to $\mathcal{O}(Nd)$ or $\mathcal{O}(Kd^2)$ rather than $\mathcal{O}(N^2d)$. Empirically, we set the hidden dimension $d$ to be only 20\%--30\% that of the teacher, yielding a model up to 5$\times$ smaller.

\noindent\textbf{\textit{Compatibility with Teacher Outputs}.} The student architecture preserves the same input-output shapes as the teacher to ensure straightforward distillation. An initial convolution layer maps the multi-channel input sequence to the reduced hidden dimension $d$, feeding into the chosen lightweight backbone. Finally, a deconvolution (or reshaping) head reconstructs the forecast $\hat{Y}_s = \mathcal{G}(X)$. As detailed in Stage 3, we will align $\hat{Y}_s$ with the teacher outputs at multiple spectral bands to guide the learning of crucial high-frequency details and low-frequency trends.

\subsection{Stage 3: Offline Training with Knowledge Distillation}
After pretraining a frequency-decoupled teacher  and designing a lightweight student, we perform offline training under a knowledge distillation framework. Suppose we have one or more teacher networks, each of which has been trained with the convolution-Transformer “high-frequency vs. low-frequency” decoupling mechanism.

\subsubsection{Frequency-Aligned Distillation}
For a single teacher, our focus is on transferring the teacher’s multi-scale spectral embeddings into the student. Specifically,  produces latent representations  containing high-frequency and low-frequency components. We extract these frequency-specific features and construct the \textit{spectral transfer loss}:
\begin{equation}
\begin{split}
\mathcal{L}_\text{KD} 
= 
& \|\Psi_h(\mathcal{G}(X)) - \Psi_h(\mathcal{F}(X))\|_2^2 \\
& + \alpha \,\|\Psi_l(\mathcal{G}(X)) - \Psi_l(\mathcal{F}(X))\|_2^2,
\end{split}
\end{equation}
here, $\alpha$ balances the focus on aligning high and low frequencies. Meanwhile, we also minimize the student's loss on the forecasting target. Thus, the overall optimization objective is:
\begin{equation}
\min_{\theta_{\mathcal{G}}}
\Bigl(
  \|\mathcal{G}(X) - Y\|_2^2
  \;+\;
  \lambda \,\mathcal{L}_\text{KD}
\Bigr),
\end{equation}
here, $\lambda$ is a hyperparameter that balances the forecasting error and the distillation loss.

\subsubsection{Multi-Teacher Distillation}
We can have multiple teacher models $\{\mathcal{F}_1, \ldots, \mathcal{F}_M\}$, each excelling in different spatial/frequency domains or physical environments. Simply averaging all teacher outputs or features may lead to information loss~\cite{yang2025multi} when conflicts or differences exist among teachers. To better leverage their complementarity, we adopt the \textit{Agree to Disagree (A2D)} method proposed by Du et al.~\cite{du2020agree}, treating multi-teacher distillation as a small-scale \textbf{multi-objective optimization} problem. In each iteration, we solve a subproblem in the gradient space to adaptively weight the teacher gradients.

\noindent\textbf{(1) Define the distillation loss for each teacher.} For the $m$th teacher $\mathcal{F}_m$, define its frequency alignment distillation loss as:
\begin{equation}
   \ell_m(\theta_{\mathcal{G}})
   \;=\;
   \|\mathcal{G}(X) - Y\|^2
   \;+\;
   \lambda \,\|\Psi(\mathcal{G}(X)) - \Psi(\mathcal{F}_m(X))\|^2,
\end{equation}
Here, $\Psi(\cdot)$ extracts features and aligns them in the frequency domain. Each teacher provides a loss signal $\ell_m$.

\noindent\textbf{(2) Perform multi-objective optimization in the gradient space.} A2D solves the teacher gradient weighting coefficients $\{\alpha_m\}$ on the "capped simplex"~\cite{du2020agree} during each mini-batch update using the following equation:
\begin{equation}
\begin{split}
   \min_{\{\alpha_m\}} & \quad
   \frac{1}{2}
   \Bigl\|
      \sum_{m=1}^M \alpha_m \,\nabla_{\theta}\ell_m
   \Bigr\|^2 \\
   \text{s.t.} & \quad
   \sum_{m=1}^M \alpha_m = 1, \quad
   0 \,\le\, \alpha_m \,\le\, C.
\end{split}
\end{equation}
Here, $C \in (0,1]$ is a hyperparameter that controls the tolerance for divergence among teachers.

\noindent\textbf{(3) Update the student network parameters.} Let $\alpha_m^*$ be the optimal solution obtained from the above equation. The student updates its parameters using the following gradient descent direction:
 \begin{equation}
   d
   \;=\;
   -\sum_{m=1}^M
     \alpha_m^* \,\nabla_{\theta}\ell_m.
\end{equation}
Let the learning rate be $\eta$, then the student updates its parameters as:
\begin{equation}
   \theta_{\mathcal{G}}
   \;\leftarrow\;
   \theta_{\mathcal{G}}
   \;-\;
   \eta \,d.
\end{equation}
In this process, each teacher's influence on the student is dynamically determined by $\alpha_m^*$. If a teacher's gradient direction conflicts significantly with others, its weight decreases accordingly. This avoids "blind averaging" while effectively preserving useful information from most teachers. By introducing A2D, frequency alignment distillation in multi-teacher scenarios no longer relies on simple averaging of all teacher features or outputs. Instead, it adaptively assigns weights at the gradient level. This approach fully leverages each teacher’s strengths in modeling different frequency bands (high/low) or scenarios, improving the student model’s overall forecasting ability.

\subsection{Stage 4: Prediction with Lightweight Student}
After training, the student model $\mathcal{G}$ is deployed for online inference. Given a new input sequence $X_{\text{new}}$, the prediction is generated in a single forward pass:
\begin{equation}
\hat{Y}_{\text{new}} = \mathcal{G}(X_{\text{new}}) \in \mathbb{R}^{H \times W \times C \times \Delta}.
\end{equation}
The lightweight architecture removes the teacher's quadratic-complexity Transformer blocks and deep convolution stacks. For example, on the NS equation dataset, inference speed increases by $2.28 \times$ compared to the teacher. Importantly, frequency-aligned knowledge from distillation enables $\mathcal{G}$ to keep multi-scale modeling capability: local convolutions approximate high-frequency patterns learned from the teacher’s CNN module, and residual connections propagate global trends similar to attention mechanisms. This approach ensures accurate predictions under both steady-state and transient disturbances without explicit frequency transforms, achieving an optimal balance between efficiency and spectral fidelity (see Figure 1).

\section{Experiment}
\begin{table}[t]
\small
\centering
\caption{Detailed information for benchmarks. $N_{tr}$, $N_{ev}$ and $N_{te}$ represent the number of instances in the training, evaluation, and test sets, while $I_l$ and $O_l$ denote the lengths of the input and prediction sequences, respectively. $N_{v}$ denotes the number of variables.}
\vspace{-0.2cm}
\label{tab:dataset}
\resizebox{0.48\textwidth}{!}{%
    \begin{tabular}{c|cccccccc}
    \toprule
    \textbf{Dataset} & \textbf{$N_{tr}$} & \textbf{$N_{ev}$} & \textbf{$N_{te}$} & \textbf{$N_v$} & \textbf{Resolution} & \textbf{$I_l$} & \textbf{$O_l$} & \textbf{Interval} \\
    \midrule
    \textbf{RBC} & 35,718 & 4,465  & 4,465  & 1 & (32, 480) & 13 & 12 & 5 mins\\
    \textbf{TaxiBJ+} & 1,660   & 208   & 208   & 2 & (128, 128)& 5  & 15 & 1 day \\
    \textbf{WeatherBench} & 4,158   & 445   & 500   & 3 & (32, 64)& 6  & 36 & 1 hour \\
    \textbf{NSE} & 1,000   & 100   & 100   & 1 & (64, 64)  & 10 & 10 & 1 step\\
    \bottomrule
    \end{tabular}%
}
\vspace{-0.5cm}
\end{table}

\subsection{Experimental Settings}
\textbf{Benchmarks.} As shown in Table~\ref{tab:dataset}, we use a wide range of scenarios to evaluate our model which includes four datasets. Here are the descriptions of these datasets.
\begin{enumerate}[leftmargin=*]

\item \textbf{Rayleigh-Bénard Convection(RBC)}~\cite{wu2024neural} is calculated by the Lattice Boltzmann Method to solve the 2-d fluid thermodynamics equations for two-dimensional turbulent flow.

\item \textbf{TaxiBJ+}~\cite{wu2024earthfarsser} is a spatiotemporal traffic flow dataset derived from GPS trajectories of taxis in Beijing. It divides the city into grid regions and records inflow and outflow every 30 minutes, forming a temporal-spatial data stream. Covering a long time span (e.g., 2013–2016), it includes various travel scenarios such as weekdays, weekends, and holidays, making it a widely used benchmark for evaluating spatiotemporal forecasting models.

\item \textbf{WeatherBench}~\cite{lin2022conditional} selects 2,048 grid points ($H \times W = 32 \times 64$) on the Earth's sphere and provides hourly meteorological data. The experiment uses four variables: temperature ($K$), cloud cover ($\% \times 10^(-1)$), humidity ($\% \times 10$), and surface wind speed components ($ms^{-1}$). The data spans from January 1, 2010, to December 31, 2018. Both input and forecast time lengths are set to 12 hours.

\item \textbf{Navier-Stokes Equation(NSE)}~\cite{li2020fourier} describes the dynamics and mass transport of the general fluid. We select the two-dimensional equations for an incompressible viscous fluid with a viscosity coefficient of $10^{-5}$ to test our model for learning complicated fluid dynamics with high Reynolds numbers. 

\noindent\textbf{Teacher Models:} We use two advanced neural network architectures as teacher networks. The detailed descriptions are as follows.
\begin{itemize}[leftmargin=*]
    \item \textbf{ST-AlterNet (T$_1$)}: Our proposed teacher algorithm combines Transformer and CNN in a serial architecture. The model first extracts local spatial features through convolutional layers, then captures global temporal dependencies via multi-head self-attention mechanisms. This hybrid design enables joint modeling of short-range and long-range spatiotemporal patterns.

    \item \textbf{SimVP (T$_2$)}: As baseline teacher model, we adopt the standard SimVP architecture \cite{gao2022simvp} which implements pure convolutional operations for video prediction. Its deep hierarchical structure with temporal skip connections serves as a strong spatial-temporal feature extractor.

\end{itemize}

\noindent\textbf{Student Models:} We select three lightweight neural network architectures as student networks. The descriptions are as follows, with the core code provided in the appendix.
\begin{itemize}[leftmargin=*]
    \item \textbf{U-Net}: Selected for its encoder-decoder structure with skip connections, particularly effective in preserving fine-grained spatial details through symmetric contracting and expanding paths.    

    \item \textbf{ResNet}: Utilizes residual blocks with identity mappings to enable stable training of deep networks. The bottleneck design balances computational efficiency and feature representation capacity.

    \item \textbf{MLP-Mixer}: Adopts the isotropic architecture that processes spatial and temporal tokens through parallel MLP layers, providing a parameter-efficient baseline without inductive bias.
\end{itemize}

\end{enumerate}

\begin{table}[t]
    \centering
    \caption{Distillation performance on Weatherbench and TaxiBJ+ with MAE, PSNR, SSIM, MSE, MAE, and SSIM.}
    \resizebox{\linewidth}{!}{
    \begin{tabular}{l|ccc|ccc}
    \toprule
    \multirow{2}{*}{Method} 
    & \multicolumn{3}{c|}{Weatherbench} 
    & \multicolumn{3}{c}{TaxiBJ+} \\ 
    \cmidrule(lr){2-4} \cmidrule(lr){5-7}
    & MAE & PSNR & SSIM 
    & MSE & MAE & SSIM \\
    \midrule
    T$_{1}$: ST-AlterNet 
    &0.7287 &33.3219 &0.9493
    &0.0683 &0.1172 &0.9701 \\
    T$_{2}$: SimVP 
    &0.7475 &32.1124 &0.9331
    &0.0697 &0.1233 &0.9653 \\
    \midrule
    S: U-Net  
    &0.9822 &29.3741 &0.8635
    &0.0831 &0.1354 &0.9532 \\
    +AVER-MKD ($w$ $AB$ $loss$)
    &0.8541 &31.4527 &0.8812
    &0.0728 &0.1241 &0.9624 \\
    +AEKD ($w$ $AB$ $loss$)  
    &0.8925 &30.7843 &0.8724
    &0.0753 &0.1279 &0.9587 \\
    +AEKD ($w$ $MSE$ $loss$)
    &\textbf{0.8457} &\textbf{31.6921} &\textbf{0.8849}
    &\textbf{0.0715} &\textbf{0.1228} &\textbf{0.9639} \\
    \midrule
    S: ResNet 
    &0.9645 &30.0192 &0.8662
    &0.0876 &0.1421 &0.9483 \\
    +AEKD ($w$ $AB$ $loss$) 
    &0.9218 &30.5124 &0.8753
    &0.0802 &0.1337 &0.9549 \\
    +AEKD ($w$ $MSE$ $loss$) 
    &\textbf{0.9102} &\textbf{30.8237} &\textbf{0.8801}
    &\textbf{0.0784} &\textbf{0.1305} &\textbf{0.9572} \\
    \midrule
    S: MLP-Mixer 
    &1.1826 &24.9732 &0.7821
    &0.0953 &0.1527 &0.9375 \\
    +AEKD ($w$ $AB$ $loss$) 
    &1.1534 &25.2175 &0.7879
    &0.0918 &0.1473 &0.9402 \\
    +AEKD ($w$ $MSE$ $loss$) 
    &\textbf{1.1428} &\textbf{25.3216} &\textbf{0.7893}
    &\textbf{0.0901} &\textbf{0.1452} &\textbf{0.9427} \\
    \bottomrule
    \end{tabular}}
    \label{tab:new_datasets}
\end{table}
\begin{figure}[t]
    \centering
    \includegraphics[width=1\linewidth]{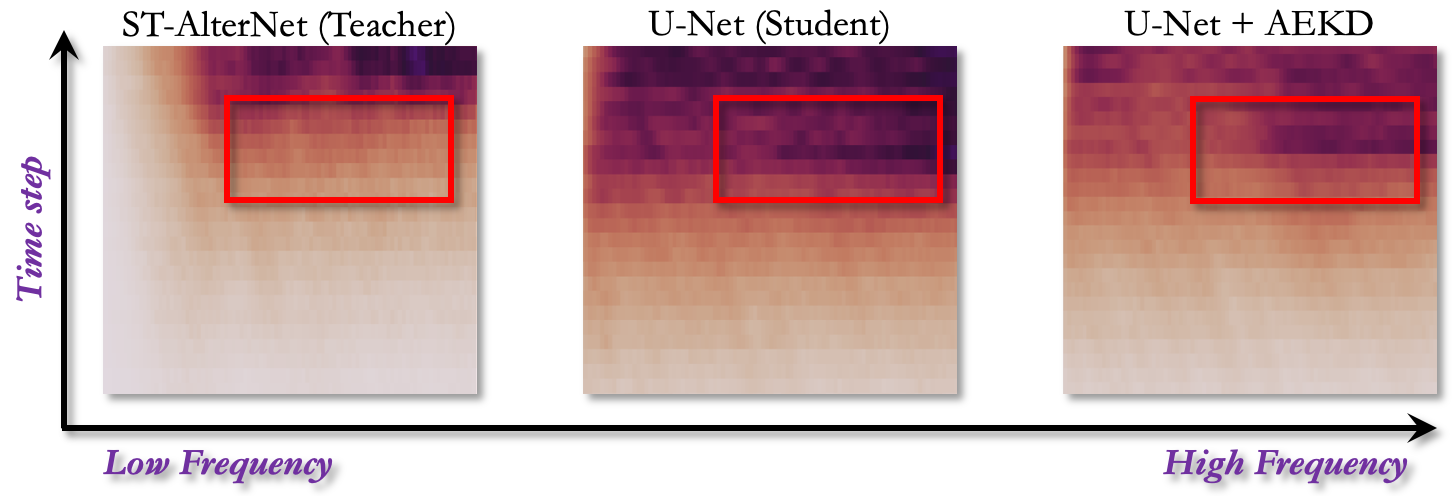}
    \caption{Spectral characteristics analysis in the Weatherbench dataset.}
    \label{fig:kd_low_high}
\vspace{-0.5cm}
\end{figure}

\begin{table}[t]
    \caption{Performance Comparison on NS Dataset with U-Net Student Model. The table presents MSE, MAE, and SSIM scores for various methods, highlighting the best-performing ones with a checkmark (green) and the worst with a cross (red). Bold indicates best performance.}
    \vspace{-0.4cm}
    \label{tab:ns_unet}
    \vskip 0.10in
    \centering
    \begin{small}
        \begin{sc}
            \setlength{\tabcolsep}{4pt}
            \begin{tabular}{@{}l|ccc|c@{}}
                \toprule
                \textbf{Method} & \textbf{MSE} & \textbf{MAE} & \textbf{SSIM} & \textbf{Overall} \\
                \midrule
                Baseline (S)    & 0.141        & 0.239        & 0.658         & -- \\
                AVER-MKD       & 0.136        & \textbf{0.233 }       & 0.670         & \textcolor{green}{\checkmark} \\
                AEKD           & \textbf{0.135  }      & 0.235        & \textbf{0.663}         & \textcolor{green}{\checkmark} \\
                CAMKD          & 0.136        & 0.234        & 0.667         & \textcolor{green}{\checkmark} \\
                \bottomrule
            \end{tabular}
        \end{sc}
    \end{small}
    \vspace{-0.5cm}
\end{table}

\begin{table}[t]
    \caption{Performance Comparison on RBC Dataset with U-Net Student Model. This table compares MSE, MAE, SSIM, and overall performance, with the best results marked by a green checkmark and the worst by a red cross. Bold indicates best performance.}
    \vspace{-0.4cm}
    \label{tab:rbc_unet}
    \vskip 0.10in
    \centering
    \begin{small}
        \begin{sc}
            \setlength{\tabcolsep}{3pt}
            \begin{tabular}{@{}l|ccc|c@{}}
                \toprule
                \textbf{Method} & \textbf{MSE} & \textbf{MAE} & \textbf{SSIM} & \textbf{Overall} \\
                \midrule
                Baseline (S)        & 0.0267      & 0.114        & 0.723         & -- \\
                AVER-MKD           & 0.0264      & 0.114        & 0.726         & \textcolor{green}{\checkmark} \\
                AEKD (w ABLoss)    & 0.0281      & 0.118        & 0.714         &$ \textcolor{red}{\times} $\\
                CAMKD              & \textbf{0.0261} & \textbf{0.112} & \textbf{0.728} & $\textcolor{green}{\checkmark}$ \\
                Single Teacher (VAN) & 0.0265    & 0.114        & 0.726         &$ \textcolor{green}{\checkmark} $\\
                \bottomrule
            \end{tabular}
        \end{sc}
    \end{small}
\end{table}

\begin{table}[t]
    \caption{Performance Comparison on RBC Dataset with ResNet Student Model. MSE, MAE, and SSIM are reported, with the best methods marked with a green checkmark and the worst with a red cross. Bold indicates best performance.}
    \vspace{-0.4cm}
    \label{tab:rbc_resnet}
    \vskip 0.10in
    \centering
    \begin{small}
        \begin{sc}
            \setlength{\tabcolsep}{3pt}
            \begin{tabular}{@{}l|ccc|c@{}}
                \toprule
                \textbf{Method} & \textbf{MSE} & \textbf{MAE} & \textbf{SSIM} & \textbf{Overall} \\
                \midrule
                Baseline (S)    & 0.0356      & 0.132        & 0.686         & -- \\
                AEKD (w ABLoss) & 0.0395      & 0.139        & 0.675         & $\textcolor{red}{\times}$ \\
                CAMKD           & \textbf{0.0331} & \textbf{0.125} & \textbf{0.706} & $\textcolor{green}{\checkmark} $\\
                \bottomrule
            \end{tabular}
        \end{sc}
    \end{small}
    \vspace{-0.5cm}
\end{table}

\noindent\textbf{Distillation Strategy.} We implement two knowledge transfer frameworks:
\begin{itemize}[leftmargin=*]
    \item \textbf{AVER-MKD}~\cite{hinton2015distilling}: A baseline method for multi-teacher knowledge distillation that guides the student model by averaging the outputs of all teachers.
    \item \textbf{AEKD}~\cite{du2020agree}: The baseline adaptive ensemble knowledge distillation method, implemented with two loss variants: \ding{182} \textit{AB Loss}~\cite{heo2019knowledge}: Adversarial boundary-aware loss that emphasizes transitional regions. \ding{183} \textit{MSE Loss}: Standard mean squared error based feature matching
\end{itemize}

\noindent\textbf{Implementation.} Based on the above experimental setup, all our experiments are conducted on two NVIDIA GeForce RTX 4090 GPUs. For the knowledge distillation part, we add AB Loss or MSE Loss in addition to the task-supervised MSE loss, forming a composite loss to better fit the teacher model's feature distribution. All models use the Adam optimizer with a fixed learning rate of $1 \times 10^{-4}$, trained for 300 epochs, and early stopping is applied based on monitoring the loss on the validation set. We use PyTorch's default parameter initialization and set the random seed to 42 to ensure reproducibility. The model weights with the best performance on the validation set are retained for the testing phase. 


\subsection{Main Results}
The Table~\ref{tab:new_datasets} shows the distillation performance on the Weatherbench and TaxiBJ+ datasets. We compare different teacher models (T$_{1}$: ST-AlterNet and T$_{2}$: SimVP) with student models. ST-AlterNet outperforms SimVP, especially on Weatherbench, with better MAE, PSNR, and SSIM values. Next, we compare student models trained with different distillation strategies like AVER-MKD and AEKD. U-Net, ResNet, and MLP-Mixer all show better performance with AEKD (especially with AB Loss or MSE Loss). For example, U-Net with AEKD (w AB Loss) achieves MAE of 0.8541, PSNR of 31.4527, and SSIM of 0.8812. ResNet and MLP-Mixer also improve in MAE and SSIM after AEKD distillation. In summary, AEKD helps student models perform better by transferring spectral features from the teacher model. It maintains a good balance between efficiency and accuracy.

\begin{figure}[t]
    \centering
    \includegraphics[width=1\linewidth]{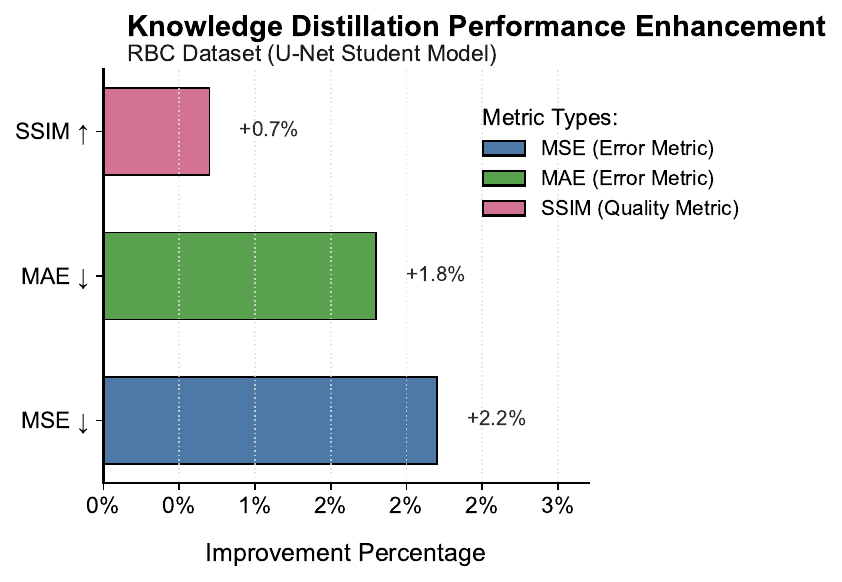}
    \caption{Performance improvement percentage on the RBC Dataset over U-Net Student Model.}
    \label{fig:enhanced_distillation}
    \vspace{-0.5cm}
\end{figure}
\subsection{Spectral characteristics analysis}

As shown in the Figure~\ref{fig:kd_low_high}, the teacher model ST-AlterNet has lower prediction errors in both low and high-frequency regions, demonstrating balanced modeling of the multi-scale evolution of meteorological systems. In contrast, the U-Net student model without distillation exhibits higher errors, especially in the high-frequency region. This suggests that the original U-Net struggles to effectively capture high-frequency information, highlighting the limitations of its simple convolutional structure in modeling cross-scale interactions—over-smoothing leads to a shift in low-frequency trends, while high-frequency transient features (e.g., thunderstorm boundaries) are distorted. After spectral decoupling distillation (SDKD), the student model’s low and high-frequency errors significantly decrease, approaching the teacher model's performance. This improvement is attributed to the two-stage distillation mechanism: \textit{(1) Low-frequency alignment}: By transferring the attention weight matrix of the teacher's Transformer module, the student model learns the global meteorological field evolution rules through its residual connections. \textit{(2) High-frequency enhancement}: Using the teacher’s latent space multi-scale features as supervision, the student model’s shallow convolution kernels focus on local perturbation patterns, thereby reducing high-frequency errors.

\subsection{Distillation Method Analysis}
As shown in Tables~\ref{tab:ns_unet}, ~\ref{tab:rbc_unet}, ~\ref{tab:rbc_resnet}, and Figure~\ref{fig:enhanced_distillation}, we compare the performance of different distillation methods on the NS and RBC datasets. The baseline method (S) performs poorly, with MSE and MAE of 0.141 and 0.239 (NS dataset), and 0.0267 and 0.114 (RBC dataset), respectively. AVER-MKD and AEKD improve performance on both datasets, with AVER-MKD reducing MSE on the RBC dataset to 0.0264, MAE to 0.114, and increasing SSIM to 0.726. However, CAMKD performs the best, with an MSE of 0.0261, MAE of 0.112, and SSIM of 0.728 on the RBC dataset, showing the best cross-scale modeling ability. CAMKD reduces high-frequency errors and improves low-frequency trend modeling accuracy through multi-teacher distillation and spectral alignment mechanisms, delivering the best predictive performance.

\subsection{Efficiency Analysis}
\begin{figure}
    \centering
    \includegraphics[width=0.8\linewidth]{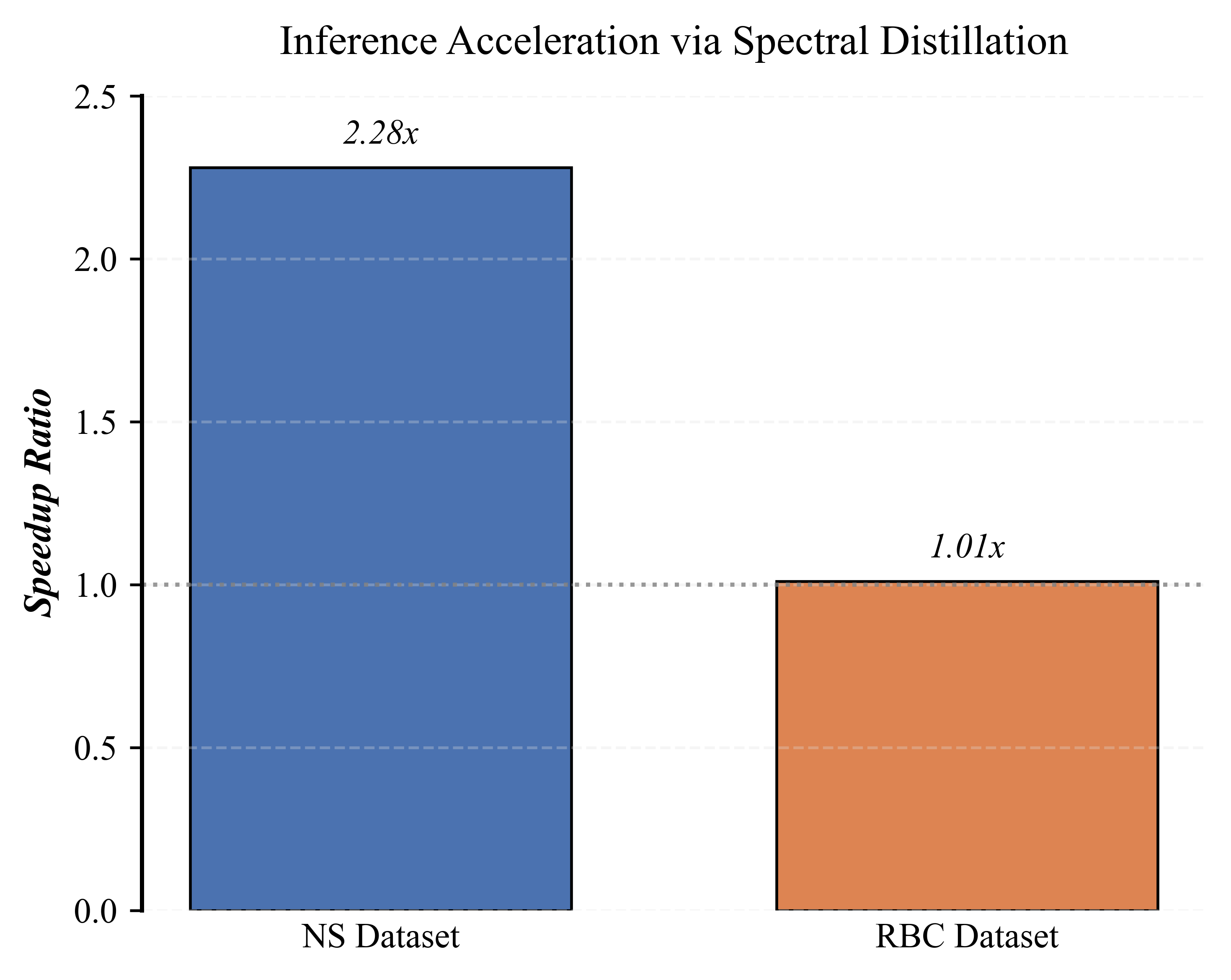}
    \caption{Inference speedup comparison between teacher and distilled student models.}
    \label{fig:kd_last}
    \vspace{-0.5cm}
\end{figure}

In this section, as shown in Figure~\ref{fig:kd_last}, we compare the inference speeds of the teacher model and the distilled student model. On the NS dataset, the teacher model (SimVP) takes 0.0115 seconds, while the distilled U-Net student model takes 0.0050 seconds, nearly halving the inference time. On the RBC dataset, the teacher model (SimVP) takes 0.0192 seconds, while the distilled U-Net student model takes 0.0189 seconds, offering a smaller yet still notable speedup. The slower speed on the RBC dataset is mainly due to its higher resolution. This indicates that while the gap on the RBC dataset is smaller, the distillation method provides a more significant acceleration on the NS dataset.
\section{Conclusion}
In summary, our proposed spectral decoupling knowledge distillation framework achieves excellent prediction performance on multiple datasets. It significantly improves the lightweight student model's ability to capture multi-scale features, including high-frequency details and low-frequency trends. It explicitly aligns the teacher model's spectral information and distills it to the student. This approach resolves the original model's limitations in cross-scale feature learning. It achieves or nearly matches the teacher model’s accuracy while reducing inference time. Our findings reveal that this spectral decoupling distillation paradigm has strong applicability and expansion potential in spatiotemporal forecasting, offering new insights and technical support for future research on multi-scale spatiotemporal modeling and deployment.

\section*{Acknowledgements}
This work is partially supported by the National Natural Science Foundation of China under Grant Number 62476264 and 62406312, the Postdoctoral Fellowship Program and China Postdoctoral Science Foundation under Grant Number BX20240385 (China National Postdoctoral Program for Innovative Talents), and the Science Foundation of the Chinese Academy of Sciences.

{
    \small
    \bibliographystyle{ieeenat_fullname}
    \bibliography{main}
}

\end{document}